\pdfoutput=1

\documentclass[11pt]{article}
\usepackage[final]{acl}
\usepackage{times}
\usepackage{latexsym}
\usepackage{multicol}
\usepackage{adjustbox}

\usepackage{footnotehyper}

\usepackage{microtype}

\usepackage{inconsolata}

\usepackage{graphicx}

\makeatletter

\patchcmd{\NAT@test}{\else \NAT@nm}{\else \NAT@nmfmt{\NAT@nm}}{}{}

\DeclareRobustCommand\citepos
  {\begingroup
   \let\NAT@nmfmt\NAT@posfmt
   \NAT@swafalse\let\NAT@ctype\z@\NAT@partrue
   \@ifstar{\NAT@fulltrue\NAT@citetp}{\NAT@fullfalse\NAT@citetp}}

\let\NAT@orig@nmfmt\NAT@nmfmt
\def\NAT@posfmt#1{\NAT@orig@nmfmt{#1's}}

\makeatother

\usepackage{forest,adjustbox}
\useforestlibrary{linguistics}
\forestapplylibrarydefaults{linguistics}

\usepackage{avm}

\title{Revisiting Supertagging for Faster HPSG Parsing}

\author{Olga Zamaraeva \and Carlos Gómez-Rodríguez \\
  Universidade da Coruña, CITIC  \\
  Departamento de Ciencias de la Computación y Tecnologías de la Información\\
  Campus de Elviña s/n, 15071, A Coruña, Spain\\
  \texttt{\{olga.zamaraeva, carlos.gomez\}@udc.es} }

\begin{document}
\maketitle

\begin{abstract}
    We present new supertaggers trained on English grammar-based treebanks and test the effects of the best tagger on parsing speed and accuracy. The treebanks are produced automatically by large manually built grammars and feature high-quality annotation based on a well-developed linguistic theory (HPSG). The English Resource Grammar treebanks include diverse and challenging test datasets, beyond the usual WSJ section 23 and Wikipedia data. 
HPSG supertagging has previously relied on MaxEnt-based models. We use SVM and neural CRF- and BERT-based methods and show that both SVM and neural supertaggers achieve considerably higher accuracy compared to the baseline and lead to an increase not only in the parsing speed but also the parser accuracy with respect to gold dependency structures. Our fine-tuned BERT-based tagger achieves 97.26\% accuracy on 950 sentences from WSJ23 and 93.88\% on the out-of-domain technical essay \emph{The Cathedral and the Bazaar} (\emph{cb})).  We present experiments with integrating the best supertagger into an HPSG parser and observe a speedup of a factor of 3 with respect to the system which uses no tagging at all, as well as large recall gains and an overall precision gain. We also compare our system to an existing integrated tagger and show that although the well-integrated tagger remains the fastest, our experimental system can be more accurate. Finally, we hope that the diverse and difficult datasets we used for evaluation will gain more popularity in the field: we show that results can differ depending on the dataset, even if it is an in-domain one. We contribute the complete datasets reformatted for Huggingface token classification.
\end{abstract}

\section{Introduction}
\label{sec:intro}

We present new supertaggers for English and use them to improve parsing efficiency for Head-driven Phrase Structure Grammars (HPSG). Grammars have been gaining relevance in the natural language processing (NLP) landscape \citep{someya2024targeted}, since it is hard to interpret and evaluate the output of NLP systems without robust theories. 

Head-Driven Phrase Structure Grammar \citep[HPSG]{Pol:Sag:94} is a theory of syntax that has been applied in computational linguistic research (see \citealt{bender:emerson:handbook} \S3-\S4). At the core of such research are precision grammars which encode a strict notion of grammaticality\,---\,their purpose is to cover and generate only grammatical structures. They include a relatively small set of phrase-structure rules and a large lexicon where lexical entries contain information about the word's syntactic behavior. HPSG treebanks (and the grammars that produce them) encode not only constituency but also dependency and semantic relations and have proven useful in natural language processing, e.g.\ in grammar coaching \citep{flickinger2013toward, da2016syntactic,da2020automated}, natural language generation \citep{hajdik2019neural}, and as training data for high precision semantic parsers \citep{lin2022towards, chen-etal-2018-accurate, buys2017robust}. Assuming a good parse ranking model, a treebank is produced automatically by parsing text with the grammar, and any updates are encoded systematically in the grammar, with no need of manual treebank annotation.\footnote{For a good parse ranking model, it is necessary to select ``gold'' parses from a potentially large parse forest at least once. This can be done semi-automatically \citep[][]{packard2015full}.}

HPSG parsing, which is typically bottom-up chart parsing, is both relatively slow and RAM-hungry. Often, more than a second is required to parse a sentence (see Table \ref{tab:speed}), and sometimes the performance is prohibitively bad for long sentences, with a typical user machine requiring unreasonable amounts of RAM to finish parsing with a large parse chart \citep{marimon2014automatic,oepen2002efficient}. It is important to emphasize that \textbf{this is the state of the art in HPSG parsing}, and its speed is one of the reasons why the true potential of HPSG parsing in NLP remains not fully realized despite the evidence that it helps create highly precise training data automatically. Approaches to speed up HPSG parsing include local ambiguity packing \citep{tomita1985efficient,malouf2000efficient,oepen2002efficient}, on the one hand, and forgoing exact search and reducing the parser search space, on the other \citep{dridan2008enhancing,dridan2009using,dridan2013ubertagging}. Here we contribute to the second line of research, aka supertagging, a technique to discard unlikely interpretations of tokens. \citet{dridan2008enhancing} and \citet{dridan2009using,dridan2013ubertagging} used maximum entropy-based models trained on a combination of gold and automatically labeled data from English, requiring large-scale computation. They report an efficiency improvement of a factor of 3 for the parser they worked with \citep{callmeier2000pet} and accuracy improvements with respect to the ParsEval metric. 

We present new models for HPSG supertagging, an SVM-based one, a neural CRF-based one, and a fine-tuned-BERT one, and compare their tagging accuracy with a MaxEnt baseline. We now have more English gold training data thanks to the HPSG grammar engineering consortium's treebanking efforts \citep{flickinger2000building,oepen2004lingo,Flickinger:11,flickinger2012deepbank}.\footnote{The data is available as part of the 2023 release of the English Resource Grammar (the ERG): \url{https://github.com/delph-in/docs/wiki/RedwoodsTop}.} It makes sense to train modern models on this wealth of gold data. Then we use the supertags to filter the parse chart at the lexical analysis stage, so that the parser has fewer possibilities to consider. We report the results of parsing all of the test data associated with the English HPSG treebanks \citep{oepen2002efficient} in comparison with parsing the same data with the same parsing algorithm but with no tagging at all, as well as with the integrated MEMM-based tagger. If we use the tagger with some exceptions, our system is the most accurate one (using the partial dependency match metric). It is not faster that the MEMM-based tagger integrated into the parser for production mode, although it is of course much faster than parsing without tagging (by a factor of 3). 


The paper is organized as follows.
In \S\ref{sec:bg}, we give the background necessary for understanding the provenance of our training data.  \S\ref{sec:methods} presents the methodology, starting from previous work (\S\ref{sec:related}). We then describe our training and evaluation data (\S\ref{sec:data}), and finally how we trained the new supertaggers (\S\ref{sec:neural}). In \S\ref{sec:results}, we present the results: first for the accuracy of the supertagger (\S\ref{sec:results-acc}) and then for the parsing experiments, including parsing speed and parsing accuracy (\S\ref{sec:parsing-speed}).  

We trained the neural models  with NVIDIA GeForce RTX 2080 GPU, CUDA version 11.2. The SVM model and the MaxEnt baseline were trained using Intel Core i7-9700K 3,60Hz CPU. The parser was run on the same CPU.  
The code and configurations for the reported results as well as the datasets are online.\footnote{\url{https://github.com/olzama/neural-supertagging}} The original data we used is publicly available.$^2$ Further details can be found in the Appendix. 

\section{Background}
\label{sec:bg}

Below we explain HPSG lexical types (\S\ref{sec:lextypes}), which serve as the tags that we predict, and in \S\ref{sec:treebanks}, we give the background on the English treebanks which served as our training and evaluation data. \S\ref{sec:hpsg-parsing} is a summary for HPSG parsing and the specific parser that we are using for the experiments.

\subsection{Lexical types}
\label{sec:lextypes}

Any HPSG grammar consists of a hierarchy of types, including phrasal and lexical types, and of a large lexicon which can be used to map surface tokens to lexical types. Each token in the text is recognized by the parser as belonging to one or more of the lexical entries in the lexicon (assuming such an orthographic form is present at all). Lexical entries, in turn, belong to lexical types (Figure \ref{fig:hier}). 
\begin{figure}[h!]
    \centering
    \small
    \begin{forest}
    [\emph{sign}[\emph{word} [\emph{verb}\\\begin{avm}
\[LOCAL\|HEAD \emph{verb} \]
\end{avm} [\emph{main verb}\\\begin{avm}
\[NONLOCAL\|QUE \< \> \]
\end{avm}[bark-v1[\emph{bark}]]]][\emph{noun}\\\begin{avm}
\[LOCAL\|HEAD \emph{noun} \]
\end{avm} [\emph{mass-count noun}\\\begin{avm}
\[INDEX\|DIV + \]
\end{avm}[bark-n1[\emph{bark}]]]]]]
    \end{forest}
    \caption{Part of the HPSG type hierarchy (simplified; adapted from ERG). NB: This is not a derivation.}
    \label{fig:hier}
\end{figure}
Lexical types are similar to POS tags but are more fine grained (e.g.\ a precision grammar may distinguish between multiple types of proper nouns or multiple types of \emph{wh}-words, etc). Figure \ref{fig:hier} shows the ancestry of two senses of the English word \emph{bark}, a verb (to bark) and a noun (tree bark). The types differ from each other in features and their values. For example, the HEAD feature value is different for nouns and verbs; one of the characteristics of the main verb type is that it is not a question word; the noun subtype denotes divisible entities, etc. The token \emph{bark} will be interpreted as either a verb or a noun during lexical analysis parsing stage. After the lexical analysis, the bottom-up parser runs a constraint unification-based algorithm \citep{carpenter1992logic} to return a (possibly empty) set of parses. To emphasize, a parser in this context is a separate program implementing a parsing algorithm. The grammar is the type hierarchy which the parser takes as input along with the sentence to parse.


\subsection{The ERG treebanks}
\label{sec:treebanks}
The English Resource Grammar \citep[ERG;][]{flickinger2000building,Flickinger:11} is a broad-coverage precision grammar of English implemented in the HPSG formalism. The latest release is from 2023.\footnote{\url{https://github.com/delph-in/docs/wiki/ErgTop}}  Its intrinsic evaluation relies on a set of English text corpora. Each release of the ERG includes a treebank of those texts parsed by the current version. The parses are created automatically and the gold structure is verified manually. 
Treebanking in the ERG context is the process of choosing linguistically (semantically) correct structures from the multiple trees corresponding to one string that the grammar may produce. Fast treebanking is made possible by automatically comparing parse forests and by discriminant-based bulk elimination of unwanted trees \citep{oepen1999incr,packard2015full}. The treebanks are stored as databases that can be processed with specialized software e.g.\ Pydelphin\footnote{\url{https://pydelphin.readthedocs.io/}}.

The 2023 ERG release comes with 30 treebanked corpora containing over 1.5 million tokens and 105,155 sentences. In principle, there are 43,505 different lexical types in the ERG (cf. 48 tags in the Penn Treebank POS tagset \citep[PTB;][]{ptb:1993}) however only 1299 of them are found in the training portion of the treebank. The genres include well-edited text (news, Wikipedia articles, fiction, travel brochures, and technical essays) as well as customer service emails and transcribed phone conversations. There are also constructed test suites illustrating linguistic phenomena such as raising and control. The ERG treebanks present more challenging test data compared to the conventional WSJ23 (which is also included). The ERG 2023's average accuracy (correct structure) over all the corpora is 93.77\%; the raw coverage (some structure) is 96.96\%.  The ERG uses PTB-style punctuation tokens and includes PTB POS tags in all tokens, along with a lexical type (\S\ref{sec:lextypes}).


\subsection{HPSG parsing}
\label{sec:hpsg-parsing}

Several parsers for different variations of the HPSG formalism exist. We work with the DELPH-IN formalism \citep{Copestake:02:CLE} which is deliberately restricted for theoretical and performance considerations; it only encodes the unification operation natively (and not e.g.\ relational constraints). Still, the parsing algorithms' worst-case complexity is intractable \cite{oepen2002efficient}.
\citet[][\S{3.2.3}]{carroll:1993} (cited in \citealt[][p.1109]{bender:emerson:handbook}) states that the worst-case parsing time for HPSG feature structures is proportional to $C^2n^{\rho + 1}$ where $\rho$ is the maximum number of children in a phrase structure rule and C is the (potentially large) maximum number of feature structures. 
The unification operator takes two feature structures as input and outputs one feature structure which satisfies the constraints encoded in both inputs. Given the complex nature of such structures, implementing a fast unification parser is a hard problem. As it is, the existing parsers may take prohibitively long to parse a long sentence (see e.g.\ \citealt{marimon2014automatic} as well as \S\ref{sec:parsing-speed} of this paper).

\section{Methodology}
\label{sec:methods}

\begin{figure}
    \centering
    \includegraphics[width=0.45\linewidth]{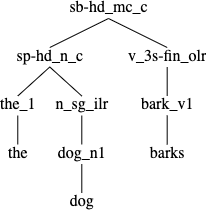}
    \includegraphics[width=0.45\linewidth]{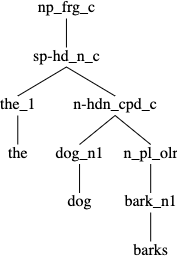}
    \caption{Two interpretations of the sentence \emph{The dog barks.} The second one is an unlikely noun phrase fragment, which would be discarded with the supertagging technique. (Trees provided by the English Resource Grammar Delphin-viz online demo.)}
    \label{fig:dog-barks}
\end{figure}

Supertagging \citep{bangalore1999supertagging} reduces the parser search space by discarding the less likely interpretations of an orthography. For example, the word \emph{bark} in English can be a verb or a noun, and in \emph{The dog barks} it is a lot less likely to be a noun than a verb (see also Figure \ref{fig:hier}). In principle, there are at least two possible interpretations of the sentence \emph{The dog barks}, as can be seen in Figure \ref{fig:dog-barks}. With supertagging, the pragmatically unlikely second interpretation would be discarded by discarding the noun lexical type (\emph{mass-count noun} in Figure \ref{fig:hier}) possibility for the word \emph{barks}.
In HPSG, there are fine-grained lexical types within the POS class (e.g.\ subtypes of common nouns or \emph{wh}-words), so the search space can be reduced further. 

In precision grammars, supertagging comes at a cost to coverage and accuracy; selecting a wrong lexical type even for one word means the entire sentence will likely not be parsed correctly. Thus the accuracy of the tagger is crucial. Related to this is the matter of how many possibilities to consider for supertags: the more are considered, the slower the parsing, but the higher the accuracy. In this paper, we experiment with a single, highest-scored tag for each token. However, we combine this strategy (which prioritizes parsing speed) with a list of tokens exempt from supertagging (which increases accuracy). 

\subsection{Previous and related work}
\label{sec:related}
 \citet{bangalore1999supertagging} introduced the concept of supertagging.  \citet{clark2003log} showed mathematically that supertagging improves parsing efficiency for a lexicalized formalism (CCG). They used a maximum entropy model; \citet{xu2015ccg} introduced a neural supertagger for CCG. \citet{vaswani2016supertagging} and \citet{tian2020supertagging} further improved the accuracy of neural-based CCG supertagging achieving an accuracy of 96.25\% on WSJ23. \citet{liu2021generating} use finer categories within the CCG tagset and report 95.5\% accuracy on in-domain test data and 81\% and 92.4\% accuracy on two out-of-domain datasets (Bioinfer and Wikipedia). \citet{prange2021supertagging} have started exploring the long-tail phenomena related to supertagging and strategies to not discard rare tags. \citet{kogkalidis2023geometry} have shown how supertagging, through its relation to underlying grammar principles, improves neural networks' abilities to deal with rare (``out-of-vocabulary'') words.\footnote{These works do not report experiments on parsing speed; they are concerned with tagging accuracy issues only.}


Supertagging experiments with HPSG parsing speed using hand-engineered grammars 
are summarized in Table \ref{tab:supertagging-effects}. In addition, there were experiments on the use of supertagging for parse ranking with statistically derived HPSG-like grammars \citep{ninomiya2007log, matsuzaki2007efficient, miyao2008feature, zhang2009hpsg, zhang2010simple, zhang2011large, zhang2012structure}. These statistically derived systems are principally different from the ERG as they do not represent HPSG theory as understood by syntacticians. In the context of the ERG, \citealt{dridan2008enhancing}  represents our baseline SOTA for the tagger accuracy. \citealt{dridan2013ubertagging} is a related work on ``ubertagging'', which includes multi-word expressions. Specifically, an ubertagger considers various multi-word spans, whereas a supertagger relies on a standard tokenizer. We use the ubertagger that was implemented for the ACE parser for the parsing speed experiments, as the baseline (\S\ref{sec:parsing-speed}). \citepos{dridan2013ubertagging} parsing accuracy results, however, are not comparable to ours; she used a different dataset, a different parser, and a different accuracy metric.

\subsection{Data}
\label{sec:data}
We train and evaluate our taggers, both for the baseline (\S\ref{sec:baseline}) and for the experiment (\S\ref{sec:neural}), on gold lexical types from the ERG 2023 release (\S\ref{sec:treebanks}). We use the train-dev-test split recommended in the release.\footnote{Download redwoods.xls from the ERG repository for details and see \url{https://github.com/delph-in/docs/wiki/RedwoodsTop}. This split is different than in \citealt{dridan2009using}.} There are 84,894 sentences in the training data, 2,045 in dev, and 7,918 in test. WSJ section 23 is used as test data, as is traditional, but so are a number of other corpora, notably \emph{The Cathedral and the Bazaar} \citep{raymond1999cathedral}, a technical essay which serves as the out-of-domain test data. 
See Table \ref{tab:acc} for the details about the test data. The column titled ``training tokens'' shows the number of tokens for the training dataset which is from the same domain as the test dataset in the row. For example, WSJ23 has 23K tokens and WSJ1-22 have 960K tokens in the ERG treebanks.

\begin{table*}[t!]
    \centering
    \begin{adjustbox}{width=0.8\textwidth}
    \begin{tabular}{llrrr}
     model&  grammar & training tok & tagset size & speed-up factor \\
     \hline
     N-gram \citep{prins2004reinforcing} & Alpino (Dutch) &24 mln & 1,365 & 2 \\
     HMM \citep[][p.~167]{blunsom2007structured}& ERG (English) & 113K & 615 & 8.5 \\
     MEMM   \citep[][p.~169]{dridan2009using} & ERG (English) & 158K & 676 & 12 \\
    \end{tabular}
    \end{adjustbox}
    \caption{Supertagging effects on HPSG parsing speed.}
    \label{tab:supertagging-effects}
 
   \end{table*}
   
   \begin{table*}[t!]
    \centering
    \begin{adjustbox}{width=\textwidth}
    
    \begin{tabular}{llrrr|rrrr|r}
       dataset  & description & sent & tok & train tok & MaxEnt & SVM & NCRF++ & BERT & D2009 \\
       \hline
        cb & technical essay & 713 & 17,244 & 0 &88.96 &89.53 &91.94 & \textbf{93.88} & 74.61\\ 
        ecpr & e-commerce & 1,088&11,550 &24,934 & 91.80& 91.99&95.09&\textbf{96.09}\\
        jh*,tg*,ps*, ron* & travel brochures & 2,116&34,098 &147,166 &90.45 &91.21 &95.44&\textbf{96.11} & 91.47\\
        petet & textual entailment & 581 &7,135 &1,578 &92.88 &95.31 &96.93&\textbf{97.71}\\
        vm32 & phone conv.\ & 1,000 & 8,730&86,630 &93.57 &94.29 & 95.62&\textbf{96.64}\\
        ws213-214 & Wikipedia & 1,470 &29,697 &161,623 & 91.31 &92.02 & 93.66&\textbf{95.59}\\
        wsj23  & Wall Street J.\ & 950& 22,987&959,709 & 94.27 & 94.72 & 96.05& \textbf{97.26}\\
        \hline
        all & all test sets as one & 7,918 & 131,441 & 1,381,645 & 91.57 & 92.28 & 94.46&\textbf{96.02}\\
        all & average & 7,918 & 131,441 & 1,381,645 &91.89 & 92.72 & 94.96& \textbf{96.18}\\
        \hline
        \hline 
        speed (sen/sec) & average & 7,918 & 131,441 & 1,381,645 & 1,024 & \textbf{7,414} & 125 & 346
        \end{tabular}
    \end{adjustbox}
    
    \caption{Baseline (MaxEnt) and experimental supertaggers' accuracy and speed on test data; tagset size is 1,299.}
    \label{tab:acc}
\end{table*}

\begin{table*}[t!]
    \centering
    \begin{adjustbox}{width=0.7\textwidth}
    \begin{tabular}{ll|ll|ll}
        model  & top mistaken token & \multicolumn{2}{c|}{top underpredicted } & \multicolumn{2}{c}{top overpredicted }  \\
         &  & all & not closely rel & all & not closely rel \\
         \hline
         BERT & \emph{i}  &  n-pn& adj-i & n-pn-gen & d-poss-my  \\
         NCRF++  &\emph{to}   & n-c  & adj-i  &v-np  & adj-i  \\
        SVM &  \emph{to}    & n-pn  & v-np*  &n-pn-gen  & adj-i  \\
        MaxEnt & \emph{have}  & n-pn & v-np*  & n-pn-gen  & adj-i 
    \end{tabular}
    \end{adjustbox}
    \caption{A summary of taggers' errors}
    \label{tab:ea}
\end{table*}


\subsection{SVM, LSTM+CRF, and fine-tuned BERT}
\label{sec:neural} 
We train a liblinear SVM model with default parameters (L2 Squared Hinge loss, C=1, one-v-rest, up to 1,000 training iterations) using the scikit-learn library \citep{scikit-learn}. To train an LSTM sequence labeling model, we use the NCRF++ library \citep{yang2018ncrf}. We choose the model by training and validating 31 models up to 100 iterations with the starting learning rate of 0.009 and the batch size of 3 (the latter parameters are the largest that are feasible for the combination of our data and the library code). The best NCRF++ model is described in the Appendix in Table \ref{tab:hyper}. 
To fine-tune BERT, we use the Huggingface transformers library \citep{wolf2019huggingface} and Pytorch \citep{paszke2017automatic}. We try both `base-bert-cased' and `base-bert-uncased' pretrained models which we fine-tune for up to 50 epochs (stopping once there is no improvement for 5 epochs) with weight decay=0.01. The `cased' model with learning rate 2e-5 achieves the best dev accuracy (Table \ref{tab:dev-neural}).

We construct feature vectors similarly to what is described in \citealt{dridan2009using} and ultimately in \citealt{ratnaparkhi1996maximum}. The training vector consists of the word orthography itself, the two previous and the two subsequent words, the word's POS tag, and, for autoregressive models, the two gold lexical type labels for the two previous words. Nonautoregressive models simply do not have the previous tag features. The test vector is the same except, for autoregressive models, instead of the gold labels for the two previous tokens, it has labels assigned to the two previous tokens by the model itself in the previous evaluation steps (an autoregressive model). The word orthographic forms come from the treebank derivation terminals obtained using the Pydelphin library.\footnote{\url{https://pydelphin.readthedocs.io/}} The PTB-style POS tags come from the treebanks and they were automatically assigned by an HMM-based tagger that is part of the ACE parser code. The POS tags provided by the parser are per token, not per terminal, so for terminals which consist of more than one token, we map the combination of more than one tag to a single PTB-style tag using a mapping constructed manually by the first author for the training data. Any combination of tags not in the training data are at test time mapped to the first tag based on that being the most frequently correct prediction in the training data.\footnote{The first tag is the correct tag in about 1/3 of the cases.} We only saw 15 unknown combinations of tags in the entire dev and test data.

\subsection{The ACE HPSG Parser}
\label{sec:method:parser}
 We work with ACE \citep{crysmann2012towards}, which has seen regular releases since the publication date and remains the state-of-the-art HPSG parser. It is intended for settings which include individual use, including with limited RAM. This parser has default RAM settings\footnote{1.2GB for chart building plus 1.5 for ``unpacking'', which is a lexical disambiguation procedure.} which can be modified, and also an in-built ``ubertagger''. While the ubertagger is based on \citealt{dridan2013ubertagging}, it is not the same thing and its performance has never been published before. In particular, its tagging accuracy is unknown and we did not seek to evaluate it (evaluating a different MaxEnt model instead). The ubertagger was integrated into the ACE parser code with great care, optimizing for performance. We also do not seek to compete with such optimizations in our experiments. For our experiments, we provide ACE with the tags predicted by the best supertagger (the BERT-based supertagger) along with the character spans corresponding to the token for which the tag was predicted.\footnote{The speed of the tagging itself is negligible because the tagger tags 346 sentences per second  (0.003 sec/sen) while HPSG parsing is an order of magnitude slower.} We then prune all lexical chart edges which correspond to this token span but do not have the predicted lexical type. As such, we follow the general idea of using supertagging for reducing the lexical chart size but we do not use the same code that the integrated ubertagger uses for this procedure. We assume that our code could be further optimized for production. 

 \subsection{Exceptions for supertagging}
 As already mentioned, mistakes in supertagging are very costly for precision grammar parsing; one wrongly predicted lexical type means the entire sentence will not be parsed correctly. After the maxent-based supertaggers were trained by \citealt{dridan2009using} and \citealt{dridan2013ubertagging}, the developer of the English Resource Grammar \citeauthor{flickinger2000building} experimented with them and has come up with a list of lexical types which the supertagger tended to predict wrong. The list included fine-grained lexical types representing words such as \emph{do}, \emph{many}, \emph{less}, \emph{hard} (among many others).\footnote{The full list can be found in the release of the ERG in the folder titled `ut' (ubertagging).} Using such exception lists counteracts the effects of supertagging and slows down the parsing, while increasing accuracy. We include this exception list methodology into our experiments, but we compile our own list based on the top mistakes our supertaggers made on the dev data. 

\section{Results}
\label{sec:results}
\subsection{Tagger accuracy and tagging speed}
\label{sec:results-acc}
\subsubsection{Tagging accuracy baseline}
\label{sec:baseline}

For our baseline, we use a MaxEnt model similar to  \citealt{dridan2009using}.
While \citet{dridan2009using} used off-the-shelf TnT \citep{brants2000tnt} and C\&C  \citep{clark2003log} taggers, we use the off-the-shelf logistic regression library from scikit-learn \citep{scikit-learn} which is a popular off-the-shelf tool for classic machine learning algorithms. The baseline tagger accuracy is included in Table \ref{tab:acc}.
The details on how the best baseline model was chosen are in Appendix A. 
The results are presented in Table \ref{tab:acc}. 

\subsubsection{Tagger accuracy results}
Table \ref{tab:acc} shows that the baseline models achieve similar performance to \citealt{dridan2009using} (D2009 in Table \ref{tab:acc}) on in-domain data and are better on out-of-domain data. 
This may indicate that these models are close to their maximum performance on in-domain data on this task but adding more training data still helps for out-of-domain data.
\citepos{dridan2009using} models were trained on a subset of our data. 
\citet[][p.84]{dridan2009using} reports getting 91.47\% accuracy on the in-domain data (which loosely corresponds to row `jh*, tg*, ps*') using the TnT tagger \citep{brants2000tnt}.

The SVM and the neural models are better than the baseline models on all test datasets, and fine-tuned BERT is the best overall. On the portion of WSJ23 for which we have gold data, fine-tuned BERT achieves 97.26\%. 
 The neural models are slower than the baseline models (using GPU for decoding); on the other hand, SVM is remarkably fast  (at over 7000 sen/sec).

All models make roughly the same mistakes (Table \ref{tab:ea}), with prepositions, pronouns, and auxiliary verbs being the most misclassified tokens, and the proper noun being the least accurate tag.\footnote{In Table \ref{tab:ea}, the ``not closely related'' column represents mistakes where the true label and the predicted label differ in their general subcategory; in this column, we did not count nouns mistaken for other types of nouns, etc. We use the ERG lexical type naming convention to filter the errors. The ``n-c'' type is a subtype of common noun; the ``n-pn'' and ``n-pn-gen'' types are subtypes of proper nouns; ``v-np*'' is a subtype of verbs that take clausal complements; ``adj-i'' is a subtype of intersective adjectives, ``d-poss'' is a possessive determiner.} 

\subsection{Results: Parsing Speed and Accuracy}
\label{sec:parsing-speed}
\begin{figure}[h!]
    \centering
    \begin{adjustbox}{width=0.5\textwidth}
        
    \includegraphics{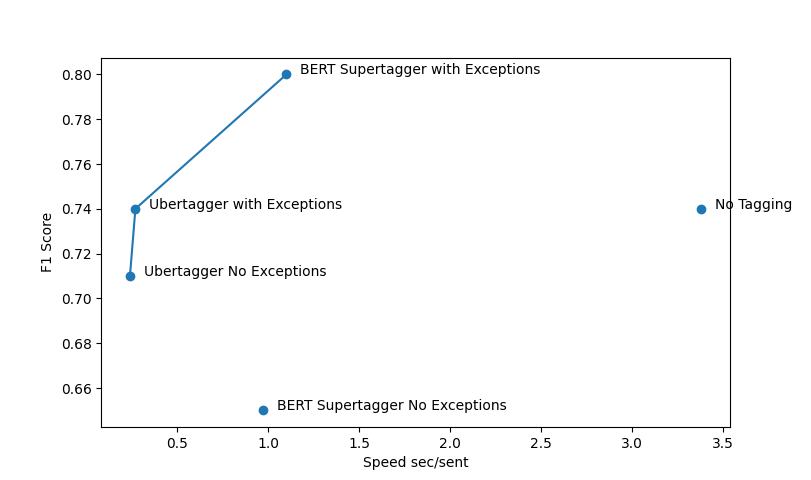}
        \end{adjustbox}
    \caption{Pareto Frontier (Speed and F-score)}
    \label{fig:pareto}
\end{figure}

We measure the effect of supertagging on parsing speed and accuracy using the ACE parser (\S\ref{sec:method:parser}). Recall that HPSG parsing is chart parsing, and for a large grammar, the charts can be huge. The goal of supertagging is to reduce the size of the lexical chart. This can make parsing faster, however if a good lexical analysis is thrown out by mistake (due to a wrong tag), the entire sentence is likely to be lost (not parsed or parsed in a meaningless way). The parser speed and the parser accuracy are therefore in tension: the more time we give the parser the more chances it will have to build the correct structure in a bigger chart. For accuracy, we report two metrics: exact match with the gold semantic structure (MRS) and partial match Elementary Dependency Match metric \citep[EDM;][]{dridan-oepen-2011-parser}. The exact match is less important because it usually can only be achieved on short, easy sentences. The EDM (and similar) is the usual practice. The results are presented in Tables \ref{tab:speed-default}-\ref{tab:parsing-acc-exact}, which are also summarized in Figure \ref{fig:pareto}.

\subsubsection{Baseline}
We compare our system with two systems: ACE with no tagging at all and ACE with the in-built ``ubertagger''. The system with no tagging at all is the baseline for parsing speed and, theoretically, the upper boundary for the parsing accuracy (as the parser could have access to the full lexical chart). However, in practice it is difficult to obtain this upper bound because it requires at least 54GB of RAM (see \S\ref{sec:ram}) and the parsing takes unreasonably long (up to several minutes per sentence). With realistic settings, the system with no tagging fails to parse some of the longer sentences because the lexical chart exceeds the RAM limit. It is precisely the problem that ubertagging/supertagging is supposed to solve: reduce the size of the lexical chart so that the parsing can be done with realistic RAM allocation and in reasonable time.

The ubertagger is a MEMM tagger based on \citealt{dridan2013ubertagging}. It was trained on millions of sentences using large computational resources (the Titan system at University of Oslo) and as such is not easily reproducible. In contrast, our BERT-based model is fairly easy to fine-tune and reproduce on an individual machine. For the purposes of parsing accuracy and speed, rather than comparing our system to other experimental taggers presented in \S\ref{sec:results-acc}, we compare it to the ubertagger because the ubertagger is integrated into the ACE parser for production and as such is a more challenging baseline. 

Below we present the results in two settings: (1) default settings, and (2) default RAM with tag exceptions. In Tables \ref{tab:speed-default} and \ref{tab:speed}, the best result is bolded, and the experimental result is italicized in the cases where it is not the best but much closer to the ubertagger than to the no-tagging baseline. 

 \begin{table*}[t!]
    \centering
    \begin{adjustbox}{width=0.7\textwidth}
    
    \begin{tabular}{llrr|ccc}
        &&&&\multicolumn{3}{c}{sec/sen}\\
       dataset  & description & sent & tok & No tagging & Ubertagging & BERT-based supertags \\
       \hline
        cb & technical essay & 713 & 17,244 & 6.15 &\textbf{0.42} &\emph{0.76} \\ 
        ecpr & e-commerce & 1,088&11,550 &0.55& \textbf{0.05}&0.52\\
        jh*,tg*,ps*, ron* & travel brochures & 2,116&34,098 &2.40&\textbf{0.13}&0.32\\
        petet & textual entailment & 581 &7,135 &1.93&\textbf{0.10}&\emph{0.23}\\
        vm32 & phone conv.\ & 1,000 & 8,730&0.73&\textbf{0.04}&\emph{0.06}\\
        ws214 & Wikipedia & 598 &12,395 &5.68&\textbf{0.42}&\emph{1.24}\\
        wsj23  & Wall Street J.\ & 950& 22,987&6.27&\textbf{0.46}&3.60\\  
         \hline
        all & average & 7,918 & 131,441 & 3.38&\textbf{0.24}&0.97
        \end{tabular}
    \end{adjustbox}
    
    \caption{Effects of supertagging on DEFAULT parsing speed (ACE Parser)}
    \label{tab:speed-default}
\end{table*}

   \begin{table*}[t!]
    \centering
    \begin{adjustbox}{width=0.7\textwidth}
    
    \begin{tabular}{llrr|lll|lll|lll}
       dataset  & description & sent & tok & \multicolumn{3}{c|}{No tagging} & \multicolumn{3}{c|}{Ubertagging} & \multicolumn{3}{c}{BERT-based supertags} \\
       &&&&Precision & Recall &F1&Precision & Recall &F1&Precision & Recall &F1\\
       \hline
        cb & technical essay & 713 & 17,244 & 0.89 &\textbf{0.40} &\textbf{0.55} &0.86&0.39&0.53&\textbf{0.91}&0.27&0.41\\ 
        ecpr & e-commerce & 1,088&11,550 &\textbf{0.95}& \textbf{0.93}&\textbf{0.94}&0.93&0.67&0.78&0.94&0.88&0.91\\
        jh*,tg*,ps*, ron* & travel brochures & 2,116&34,098 &\textbf{0.91}&\textbf{0.74}&\textbf{0.81}&0.89&0.61&0.71&0.91&0.69&0.78\\
        petet & textual entailment & 581 &7,135 &0.93&\textbf{0.78}&\textbf{0.85}&0.93&0.67&0.78&\textbf{0.94}&0.47&0.63\\
        vm32 & phone conv.\ & 1,000 & 8,730&0.92 & \textbf{0.88} & \textbf{0.90} & \textbf{0.93} & 0.75 & 0.83 & 0.94 & 0.52 & 0.67\\
        ws214 & Wikipedia & 598 &12,395 & 0.89&0.45&0.60&0.88&\textbf{0.54}&\textbf{0.66}&\textbf{0.92}&0.42&0.58\\
        wsj23  & Wall Street J.\ & 950& 22,987&\textbf{0.92}&0.38&0.54&0.90&\textbf{0.49}&\textbf{0.64}&0.91&0.44&0.60\\
        \hline
        all & average & 7,918 & 131,441 &0.92 &\textbf{0.65}&\textbf{0.74}&0.90&0.59&0.71&\textbf{0.93}&0.53&0.65\\
    
        \end{tabular}
    \end{adjustbox}
    
    \caption{Effects of supertagging on DEFAULT parsing accuracy (EDM metric)}
    \label{tab:parsing-acc-default}
\end{table*}

   \begin{table*}[t!]
    \centering
    \begin{adjustbox}{width=0.7\textwidth}
    
    \begin{tabular}{llrr|ccc}
        &&&&\multicolumn{3}{c}{exact match}\\
       dataset  & description & sent & tok & No tagging & Ubertagging & BERT-based supertags \\
       \hline
        cb & technical essay & 713 & 17,244 &0.13  &\textbf{0.15} &0.10 \\ 
        ecpr & e-commerce & 1,088&11,550 &\textbf{0.50}&0.47 &0.47\\
        jh*,tg*,ps*, ron* & travel brochures & 2,116&34,098 &\textbf{0.33}&0.31&0.29\\
        petet & textual entailment & 581 &7,135 &0.46&\textbf{0.52}&0.23\\
        vm32 & phone conv.\ & 1,000 & 8,730&0.55&\textbf{0.58}&0.49\\
        ws214 & Wikipedia & 598 &12,395 &0.23&\textbf{0.25}&0.14\\
        wsj23  & Wall Street J.\ & 950& 22,987&0.13&\textbf{0.15}&0.14\\
        \hline
        all & average & 7,918 & 131,441 &0.33 &\textbf{0.35}&0.26\\
    
        \end{tabular}
    \end{adjustbox}
        \caption{Effects of supertagging on DEFAULT parsing accuracy (exact match over MRS)}
    \label{tab:parsing-acc-exact-default}

\end{table*}

\subsubsection{Default parsing}

Tables \ref{tab:speed-default}, \ref{tab:parsing-acc-default}, and \ref{tab:parsing-acc-exact-default} present the results for the ACE parser default RAM limit setting (1200MB). On the ubertagger and the supertagger side, we use all the predictions and do not exclude any tags from the pruning process.

The results show that while we can parse faster with tagging (the ubertagger being the fastest), both the ubertagger and the supertagger suffer from the high cost of each tagging mistake: while the new BERT-based supertagger is more accurate, its accuracy is still not 100\%, and even at 99\% tagger accuracy, the likelihood of losing an entire sentence due to one incorrect tag is high. \citet{dridan2013ubertagging} comments on this, too, and suggests taking into account the top mistakes that the tagger makes to achieve higher recall. This is what we do below. 

   \begin{table*}[h!]
    \centering
    \begin{adjustbox}{width=0.8\textwidth}
    
    \begin{tabular}{llrr|ccc}
        &&&&\multicolumn{3}{c}{sec/sen}\\
       dataset  & description & sent & tok & No tagging & Ubertagging & BERT-based supertags \\
       \hline
        cb & technical essay & 713 & 17,244 & 6.15 &\textbf{0.46} &\emph{0.91} \\ 
        ecpr & e-commerce & 1,088&11,550 &0.55& \textbf{0.05}&0.52\\
        jh*,tg*,ps*, ron* & travel brochures & 2,116&34,098 &2.36&\textbf{0.16}&\emph{0.40}\\
        petet & textual entailment & 581 &7,135 &1.93&\textbf{0.11}&\emph{0.27}\\
        vm32 & phone conv.\ & 1,000 & 8,730&0.73&\textbf{0.05}&\emph{0.08}\\
        ws214 & Wikipedia & 598 &12,395 &5.68&\textbf{0.46}&\emph{1.48}\\
        wsj23  & Wall Street J.\ & 950& 22,987&6.27&\textbf{0.55}&4.04\\  
        \hline
        all & average & 7,918 & 131,441 &3,38 &\textbf{0.27}&1.10\\
        \end{tabular}
    \end{adjustbox}
    
    \caption{Effects of supertagging WITH EXCEPTIONS on parsing speed (ACE Parser)}
    \label{tab:speed}
\end{table*}

   \begin{table*}[h!]
    \centering
    \begin{adjustbox}{width=\textwidth}
    
    \begin{tabular}{llrr|lll|lll|lll}
       dataset  & description & sent & tok & \multicolumn{3}{c|}{No tagging} & \multicolumn{3}{c|}{Ubertagging} & \multicolumn{3}{c}{BERT-based supertags} \\
       &&&&Precision & Recall &F1&Precision & Recall &F1&Precision & Recall &F1\\
       \hline
        cb & technical essay & 713 & 17,244 & 0.89 &0.40 &0.55 &0.87&0.43&0.58&\textbf{0.9}2&\textbf{0.49}&\textbf{0.65}\\ 
        ecpr & e-commerce & 1,088&11,550 &\textbf{0.95}& \textbf{0.93}&\textbf{0.94}&0.93&0.81&0.87&0.95&0.90&0.92\\
        jh*,tg*,ps*, ron* & travel brochures & 2,116&34,098 &0.91&0.74&0.81&0.89&0.65&0.75&\textbf{0.93}&\textbf{0.80}&\textbf{0.86}\\
        petet & textual entailment & 581 &7,135 &0.93&\textbf{0.78}&\textbf{0.85}&0.92&0.69&0.79&\textbf{0.97}&0.69&0.81\\
        vm32 & phone conv.\ & 1,000 & 8,730&0.92 & \textbf{0.88} & \textbf{0.90} & 0.93 & 0.79 & 0.86 & \textbf{0.94} & 0.85 & 0.89\\
        ws214 & Wikipedia & 598 &12,395 & 0.89&0.45&0.60&0.88&0.54&0.67&\textbf{0.93}&\textbf{0.56}&\textbf{0.70}\\
        wsj23  & Wall Street J.\ & 950& 22,987&0.92&0.38&0.54&0.89&0.53&0.67&\textbf{0.94}&\textbf{0.69}&\textbf{0.80}\\
        \hline
        all & average & 7,918 & 131,441 &0.92 &0.65&0.74&0.90&0.63&0.74&\textbf{0.94}&\textbf{0.71}&\textbf{0.80}\\
    
        \end{tabular}
    \end{adjustbox}
    
    \caption{Effects of supertagging WITH EXCEPTIONS on parsing accuracy (EDM metric)}
    \label{tab:parsing-acc}
\end{table*}

   \begin{table*}[h!]
    \centering
    \begin{adjustbox}{width=0.8\textwidth}
    
    \begin{tabular}{llrr|ccc}
        &&&&\multicolumn{3}{c}{sec/sen}\\
       dataset  & description & sent & tok & No tagging & Ubertagging & BERT-based supertags \\
       \hline
        cb & technical essay & 713 & 17,244 &0.13  &0.17 &\textbf{0.21} \\ 
        ecpr & e-commerce & 1088&11,550 &0.55&\textbf{0.56} &0.48\\
        jh*,tg*,ps*, ron* & travel brochures & 2,116&34,098 &0.33&0.32&\textbf{0.38}\\
        petet & textual entailment & 581 &7,135 &0.46&0.54&\textbf{0.66}\\
        vm32 & phone conv.\ & 1,000 & 8,730&0.55&0.61&\textbf{0.67}\\
        ws214 & Wikipedia & 598 &12,395 &0.23&\textbf{0.26}&0.22\\
        wsj23  & Wall Street J.\ & 950& 22,987&0.13&0.17&\textbf{0.27}\\
        \hline
        all & average & 7,918 & 131,441 &0.33 &0.37&\textbf{0.42}\\
    
        \end{tabular}
    \end{adjustbox}
        \caption{Effects of supertagging WITH EXCEPTIONS on parsing accuracy (exact match over MRS)}
    \label{tab:parsing-acc-exact}

\end{table*}

\subsubsection{Parsing with exceptions lists}
Tables \ref{tab:speed}-\ref{tab:parsing-acc-exact} present the results for parsing with ubertagging and supertagging with exceptions. The no-tagging system's results are the same as before; we repeat them for convenience.

We have looked at the most common mistakes in the supertags in the training data and have compiled a list of 15 tags which BERT tends to predict wrong.\footnote{The list includes: some punctuation/quotation marks, the tags for out-of-vocabulary proper names, the verb \emph{is}, the pronouns \emph{my} and \emph{me}, and types for denoting times and dates. Cf.\ Table \ref{tab:ea} which shows similar findings on the test data, which we did not take into account.} On the ubertagger side, there was already a list of exceptions. The ubertagger's exception list is a list of 1715 lexical entries (words, e.g. ``my''), whereas ours is a list of 15 lexical types (tags, e.g.\ ''d-poss-my'', which is a supertype for ``my'' in the grammar). The ubertagger's list includes some of the words that we expect would be tagged with some of our excluded types, although in principle, the two models may of course make different mistakes. We did not modify the existing ubertagger nor consulted its exceptions for our list.  From the speeds that we are seeing, we conclude that our supertagger is less aggressive than the ubertagger and excludes more words from pruning, losing more in speed but winning considerably in accuracy as a result. This is what we would expect since we exclude entire lexical types and not just individual lexical items. The goal is a balanced tradeoff between accuracy and speed. We want the supertagger to be noticeably faster than the baseline and much more accurate than the ubertagger. This is what we observe in Tables \ref{tab:speed}-\ref{tab:parsing-acc-exact}.

Because pruning the lexical chart may and often will result in wrongly sacrificing the correct lexical type for a word, we expect the recall for the tagging systems to be lower compared to the no-tagging system. On the other hand, the no-tagging system will often run out of resources and so its overall accuracy may be lower for that reason. What we see in Table \ref{tab:parsing-acc} is that our supertagging system is the most precise one on most datasets and shows large recall gains on Wikipedia, Wall Street Journal Section 23, and the technical essay data. It is strictly better than the no-tagging system on WSJ23 as well as on Wikipedia and \emph{The Cathedral and the Bazaar}, and it is strictly better  than the ubertagger across the board on the partial match EDM metric. While the recall difference is partially explained by the supertagger being less aggressive in pruning, the precision has to be due to the higher accuracy of the tagging model (BERT). On the exact match metric, the ubertagger wins on two datasets: e-commerce and Wikipedia. The supertagger wins on the rest.

Our system is strictly faster than the baseline, by a factor of 3, although on two datasets (e-commerce and WSJ) it fails to achieve a speedup factor of 2. The ubertagger is still the fastest overall, remarkably by a factor of 12, on average across all datasets. This is not too surprising because the supertagger is experimental and it is hard for it to compete with the ubertagger which was integrated into the parser for production, with the focus on performance. We believe that the supertagger could be integrated better into the parser's C code in the future. In other words, its current speed is in part a purely C engineering problem. On the other hand, clearly the exceptions list would have an effect. Since we are excluding 15 types of words from pruning, the supertagger's lexical chart is likely to be bigger than the ubertagger's. This is the expected tension between speed and accuracy that we expected to see, and our supertagger system shows overall benefits in both speed and accuracy. The only dataset on which our system is not the best in accuracy is the e-commerce (ecpr). It appears that for this type of data, tagging is the least effective; we gain a 6\% speed increase with the supertagger at the cost of 3\% F-score, while the more aggressive ubertagger parses this data very fast but at the cost of 16\% F-score. We note particularly large recall gains on the WSJ data, but this may be related to the fact that statistical systems have been overtrained on WSJ so much that the effects are seen throughout the field \citep{hovy2015tagging}.

\section{Conclusion and future work}
\label{sec:conclusion}
We used the advancements in HPSG treebanking to train more accurate supertaggers. The ERG is a major project in syntactic theory and an important resource for creating high quality semantic treebanks. It has the potential to contribute to NLP tasks that require high precision and/or interpretability including probing of the LLMs, and thus making HPSG parsing faster is strategic for NLP. We tested the new supertagging models with the state-of-the-art HPSG parser and saw improvements in parsing speed as well as accuracy. We consider the results on multiple domains, well beyond the WSJ Section 23. We show promising results but also confirm that domain remains important, and purely statistical systems are brittle and often require rule-based additions in real-life scenarios. We contribute the ERG datasets converted to huggingface transformers format intended for token classification, along with the code which can be adapted for other purposes. 


\section{Limitations}
Our paper is concerned with training supertagging models on an English HPSG treebank. The limitations therefore are associated mainly with the training of the models including neural networks, and with the building of broad-coverage grammars such as the English Resource Grammar. Crucially, while our method does not require industry-scale computational resources, training a neural classifier such as ours still requires a certain amount of training data, and this means that our method assumes that a large HPSG treebank is available for training. The availability of such a treebank, in turn, depends directly on the availability of a broad-coverage grammar. While choosing the gold trees for the treebank can be done relatively fast using treebanking tools once the grammar parsed the corpus, building a broad-coverage grammar itself requires an investment of years of expert work. At the moment, such an investment was made only for a few languages (English, Spanish, Japanese, Chinese), English being the largest one. Furthermore, the coverage of a precision grammar is never perfect and regular grammar updates are needed. A limitation related to using neural networks is that while the NCRF++ library can in principle be very efficient on some tasks (e.g.\ POS tagging), with our data and large label set it proved relatively slow, and so ideally a more efficient neural architecture may be required for future work in this direction. 

\section*{Acknowledgments}

We acknowledge the European Union's Horizon Europe Framework Programme which funded this research under the Marie Skłodowska-Curie postdoctoral fellowship grant HORIZON-MSCA-2021-PF-01 (GAUSS, grant agreement No 101063104); and the European Research Council (ERC), which has funded this research under the Horizon Europe research and innovation programme (SALSA, grant agreement No 101100615). We also acknowledge grants SCANNER-UDC (PID2020-113230RB-C21) funded by MICIU/AEI/10.13039/501100011033; GAP (PID2022-139308OA-I00) funded by MICIU/AEI/10.13039/501100011033/ and ERDF, EU; LATCHING (PID2023-147129OB-C21) funded by MICIU/AEI/10.13039/501100011033 and ERDF, EU; and TSI-100925-2023-1 funded by Ministry for Digital Transformation and Civil Service and ``NextGenerationEU'' PRTR; as well as funding by Xunta de Galicia (ED431C 2024/02), and Centro de Investigación de Galicia ``CITIC'', funded by the Xunta de Galicia through the collaboration agreement between the Consellería de Cultura, Educación, Formación Profesional e Universidades and the Galician universities for the reinforcement of the research centres of the Galician University System (CIGUS).

\bibliography{supertagging}

\appendix
\section{Appendix A}
\label{sec:appendix-a}

\subsection{Tuning ranges}

\begin{table}[h!]
\begin{adjustbox}{width=\columnwidth}
\begin{tabular}{llll}
    Parameter & value & default/tuned & range  \\
    \hline
    lstm layers & 2 & tuned & 1--4 \\
    hidden dim.\ & 800& tuned & 100--1200\\
    word embeddings & glove840B& pretrained & \\
    word emb.\ dim.\ & 300& N/A & \\
    char emb.\ dim.\ & 50 & tuned & 30--50\\
    momentum & 0 & default & \\
    dropout & 0.5 & default & \\
    l2 & 1$^{-8}$ & default & \\
    
\end{tabular}
\end{adjustbox}
\caption{NCRF++ model parameters}
\label{tab:hyper}
\end{table}

BERT \citep{devlin2019bert} was fine-tuned using transformers \citep{wolf2019huggingface} and pytorch \citep{paszke2017automatic} using 4 learning rates: 1e-5, 2e-5, 3e-5, and 5e-6. Cased and uncased pretrained BERT models were tried.

\subsection{Computational resources}
We trained the neural models with a single NVIDIA GeForce RTX 2080 GPU, CUDA version 11.2. The SVM model and the MaxEnt baseline were trained using Intel Core i7-9700K 3,60Hz CPU (using single core processing for each model). We have experimented with Stochastic Gradient Descent (SGD) optimizer along with AdaGrad, Adam, and AdaDelta. The ranges for parameter values can be found in Table \ref{tab:hyper}. The decoding time (sentences per second) for the models can be found in Table \ref{tab:acc}. The training times are presented in this Appendix in Table \ref{tab:training}. The energy costs as estimated by the Python library carbontracker \citep{anthony2020carbontracker}\footnote{\url{https://pypi.org/project/carbontracker/}} are in Table \ref{tab:energy}.

\begin{table*}[t!]
    \centering
    \begin{tabular}{lll}
        Model type & models trained for tuning & total time for all models in this row (sec)  \\
        \hline
        SVM Scikit-learn & 1 & 3664 \\
        MaxEnt Scikit-learn & 14 & 106,922\\
        NCRF++ & 31 &  955,500 (approx.)\\
        BERT & 5 & 100,000 (approx.) \\ 
    \end{tabular}
    \caption{Training times for models used to choose the best baseline and best experimental models}
    \label{tab:training}
\end{table*}

\begin{table*}[t!]
    \centering
    \begin{tabular}{lll}
        Measurement & Value NCRF++ & Value BERT   \\
        \hline
        Process used & 5.55 kWh & 3.5 kWh\\
       Carbon emissions & 1.63 kg CO2 & 5.5 kg CO2  \\
       Equivalent km driven & 13 km & 5.5 km \\ 
    \end{tabular}
    \caption{Energy cost estimate for training the final NCRF++ model in 38 epochs (31 were trained in total, number of epochs varied) and for BERT 50 epochs}
    \label{tab:energy}
\end{table*}

\subsection{Development accuracies}
The development (validation set) accuracies are presented in Tables \ref{tab:dev}, \ref{tab:dev-neural}, and \ref{tab:dev-bert}. The best models are bolded. NCRF++ has nondeterministic components, and the average dev accuracy of the best (bolded) model in Table \ref{tab:dev-neural}; the average accuracy is 95.15\%; standard deviation 0.07088.

\begin{table}[t!]
    \centering
    \begin{tabular}{ll}
        Model & dev accuracy(\%)   \\
        \hline
        multinomial L2 SAG & 91.59 \\
        multinomial L2 SAG autoreg & 91.41 \\
       OVR L2 SAG & 91.18 \\
       OVR L2 SAG autoreg & 91.27 \\
       multinomial L2 SAGA & 91.53 \\
       multinomial L2 SAGA autoreg & 88.56 \\
       OVR L2 SAGA & 91.17\\
       OVR L2 SAGA autoreg & 91.26\\
       OVR L1 SAGA & \textbf{92.17} \\
       OVR L1 SAGA autoreg & 92.12 \\
       multinomial L1 SAGA & 92.12\\
       multinomial L1 SAGA autoreg & 91.17\\
       \hline
       liblinear SVM & 92.88\\
       \hline 
    \end{tabular}
    \caption{Development (validation) set accuracies for MaxEnt and SVM}
    \label{tab:dev}
\end{table}

\begin{table*}[t!]
\begin{adjustbox}{width=\textwidth}
    \centering
    \begin{tabular}{lllllll}
        optimizer & lstm layers & char embed & hid dim & embed & epochs (up to 100)  & dev accuracy(\%)   \\
        \hline
        SGD & 1 & 30 & 200 & pretrained (NCRF++) & 17 & 92.14\\
        SGD & 4 & 30 & 200 & pretrained (NCRF++) & 23 & 93.20\\
        SGD & 1 & 30 & 200 & random & 19 & 92.82\\
        SGD & 1 & 30 & 200 & glove840 & 17 & 93.69\\
        SGD & 4 & 30 & 200 & glove840 & 23 & 94.27\\
        SGD & 1 & 30 & 250 & glove840 & 20 & 93.86\\
        SGD & 4 & 30 & 400 & glove840 & 18 & 94.68\\
        SGD & 4 & 30 & 100 & glove840 & 19 & 93.07\\
        SGD & 2 & 30 & 200 & glove840 & 20 & 94.04\\
        SGD & 4 & 30 & 1200 & glove840 & 18 & 94.74\\
        SGD & 4 & 30 & 600 & glove840 & 19 & 94.62\\
        SGD & 2 & 30 & 400 & glove840 & 13 & 94.56\\
        SGD & 4 & 30 & 1400 & glove840 & 16 & 94.68\\
        SGD & 5 & 30 & 1200 & glove840 & 12 & 94.35\\
        SGD & 4 & 50 & 1200 & glove840 & 21 & 94.66\\
        SGD & 3 & 30 & 1200 & glove840 & 21 & 94.92\\
        SGD & 2 & 30 & 1200 & glove840 & 19 & 95.01\\
        SGD & 1 & 30 & 1200 & glove840 & 16 & 94.60\\
        SGD & 3 & 30 & 1400 & glove840 & 14 & 94.65\\
        SGD & 3 & 30 & 600 & glove840 & 15 & 94.86\\
        SGD & 2 & 50 & 1200 & glove840 & 19 & 95.00\\
        SGD & 2 & 30 & 1400 & glove840 & 8 & 94.57\\
        SGD+0.3 momentum & 3 & 30 & 1200 & glove840 & 21 & 94.67\\
        SGD & 2 & 30 & 1200 & glove840 & 12 & 94.69\\
        SGD & 2 & 30 & 1000 & glove840 & 17 & 94.95\\
        \textbf{SGD} & \textbf{2} & \textbf{30} & \textbf{800} & \textbf{glove840} & 17 & \textbf{95.12}\\
        adagrad &  1 & 30 & 200 & glove840 & 17 & 92.00\\
        adagrad &  4 & 30 & 200 & glove840 & 42 & 92.42\\
        adam & 1 & 30 & 200 & glove840 & 1 & 86.85 \\
        adadelta & 4 & 30 & 200 & glove840 & stuck on 73 & did not finish\\
      \end{tabular}
       \end{adjustbox}
    \caption{Development (validation) set accuracies for neural models}
    \label{tab:dev-neural}
   
\end{table*}

\begin{table}[t!]
    \centering
    \begin{tabular}{ll}
        Model & dev accuracy(\%)   \\
        \hline
       BERT cased LR 2e-5 & \textbf{96.46} \\
       BERT cased LR 1e-5 & 96.37 \\
       BERT cased LR 3e-5 & 96.31 \\
       BERT cased LR 5e-6 & 96.34 \\
       BERT uncased LR 2e-5 & 95.97 \\
    \end{tabular}
    \caption{Development (validation) set accuracies fine-tuned BERT models}
    \label{tab:dev-bert}
\end{table}

\subsection{MaxEnt model}
Below we describe in detail how we trained the baseline MaxEnt models. 

\subsubsection{MaxEnt model selection}

Rather than comparing our experimental numbers with numbers obtained by \citet{dridan2009using},\footnote{\citet[][p.84]{dridan2009using} reports getting 91.47\% accuracy on in-domain data using the TnT off-the-shelf tagger \citep{brants2000tnt}; as a sanity check, we obtain 91.94\% using an autoregressive one-versus-rest L1 SAGA MaxEnt model trained with the scikit-learn library \citep{scikit-learn} on training and test datasets very similar to the ones used by \citet{dridan2009using}. On \emph{The Cathedral and the Bazaar} with the same setup, we obtain 73.85\% compared to \citepos{dridan2009using} 74.61\%. We attribute the slight differences to the differences between TnT and scikit-learn.} 
we create our own baseline because we want to be able to compare classic models with neural models with the same amount of training data.

We experimented with autoregressive and non-autoregressive MaxEnt models and in the end chose one MaxEnt as the baseline. 

\subsubsection{MaxEnt classifiers}
We use scikit learn Python library \citep{scikit-learn} to train the baseline MaxEnt classifiers.\footnote{\url{https://scikit-learn.org/stable/modules/generated/sklearn.linear_model.LogisticRegression.html}} The scikit learn classifiers are optimized for processing a large number of observations. For that reason, we organized our evaluation data (dev and test) so as to maximize the number of observations passed to the classifier at each step. \citepos{dridan2009using} models were autoregressive; we also implemented autoregressive baseline models, and in order to make them faster at test time, we organized the evaluation data by the word's position in the sentence. So the classifier would first process all the first words in all sentences, then all the second words, etc. For nonautoregressive models, which we also tried in order to find the best-performing baseline model, we just pass the classifier the entire list of observations in their original order. 

We choose the single baseline MaxEnt model from the following types of models, by validation on the dev set: (1) MaxEnt autoregressive models which at test time, if more than one sentence is passed to the classifier, first classify all first tokens in all sentences, then all second tokens, etc; (2) MaxEnt nonautoregressive models where the observation tokens are organized in the same way as in (1); (3) MaxEnt nonautoregressive models where tokens are not reordered in any way and are stored consecutively. The best model happens to be of type (3).

All models achieve above 91\% accuracy on the dev set. The validation (dev) data consists of one Wikipedia section and one e-commerce corpus (2267 sentences and 25,076 tokens total), with both domains represented also in the training data. Our best performing MaxEnt baseline model is a nonautoregressive `One-versus-rest' (OVR) model with L1 regularization and SAGA optimizer (92.21\% accuracy on the dev set).\footnote{We tried multinomial and OVR models, L1 and L2 regularization, and SAG and SAGA solvers on the dev set. The autoregressive models were not strictly better than the nonautoregressive ones, achieving very similar accuracies on the dev set. They were also much slower, being able to process only 2 sentences per second.}

\subsection{Parsing with more RAM}
\label{sec:ram}
To give the baseline system an opportunity to build the full lexical chart, more than 50GB RAM is required (24+30), 
according to our experiments with a subset of the WSJ training data that includes 25 sentences some of which are very long and ambiguous \citep{yuret2010semeval}, presented below in Table \ref{tab:pest}. On this dataset, even with 54GB RAM, 100\% coverage is not achieved, and the parsing speed becomes intractable (77 sec/sen).
\begin{table}[h!]
\begin{adjustbox}{width=0.5\textwidth}
    \begin{tabular}{lll}
         & 2.7GB RAM (default) & 54GB RAM \\
         \hline 
       baseline coverage  & 11/25 (44\%)& 19/25 (76\%)  \\
       baseline speed (sec/sen) & 0.77& 77 \\
    \end{tabular}
        \end{adjustbox}
    \caption{Baseline (no tagging) coverage gain with more RAM on the PETE dataset (ERG version)}
    \label{tab:pest}
\end{table}
\begin{table}[h!]
    \centering
    \begin{adjustbox}{width=0.5\textwidth}
    
    \begin{tabular}{llll|ccc}
        &\multicolumn{3}{c}{2.7GB RAM (default)}&\multicolumn{3}{c}{54GB RAM}\\
       dataset  & coverage & F-score & speed & coverage & F-score & speed \\
       \hline
       ecpr &0.99&0.94&0.54&0.99&0.95&0.68\\
       jhk &0.87&0.78&2.61&0.98&0.87&8.68\\
       petet &0.92&0.85&1.96&0.99&0.92&4.34\\
        vm32 &0.98 &0.90 &0.74&0.99&0.92&0.99 \\ 
    
        \end{tabular}
    \end{adjustbox}
        \caption{Baseline (no tagging) recall gains and speed loss with generous RAM}
    \label{tab:RAM-baseline}

\end{table}

Since spending 77 sec/sen is not viable, we did not run the full experiments with 54GB RAM. We present below a subset of experiments, showing the baseline F-score gain due to higher recall. The ubertagger and the supertagger do not end up with such large lexical charts and thus do not benefit from more RAM, so we do not repeat the results from Tables \ref{sec:parsing-speed}-\ref{tab:parsing-acc} in Table \ref{tab:RAM}.
 
The `Verbmobil' (phone conversations) and the `ecpr' (e-commerce) datasets are easy to parse fast (as we see from Table \ref{tab:speed}) and on such data, using more RAM may be justified with the baseline system, however other types of data lead to the parsing time increasing noticeably. On the travel brochures data (`jhk'), the baseline system achieves an F-score of 87\% at the cost of spending 8.68 seconds per sentence, while our supertagger achieves 86\% with only 0.78 seconds/sentence and with only 2.7GB of RAM. Figure \ref{fig:pareto} summarizes the results presented in Tables \ref{tab:speed-default}-\ref{tab:parsing-acc}, showing that if we optimize for both speed and F-score, the best models include our model and the ubertagger models.
\begin{table*}[h!]
    \centering
    \begin{adjustbox}{width=0.7\textwidth}
    
    \begin{tabular}{l|ccc|ccc|ccc}
        &\multicolumn{3}{c}{No tagging}&\multicolumn{3}{c}{Ubertagging}&\multicolumn{3}{c}{BERT supertagging}\\
       dataset  & coverage & F-score & speed & coverage & F-score & speed & coverage & F-score & speed\\
       \hline
       cb &0.86&0.77&59.3&0.58&0.58&0.66&0.63&0.64&8.58\\
       ecpr &0.99&0.95&0.68&0.96&0.87&0.06&0.97&0.93&0.68\\
       jhk &0.98&0.88&8.89&0.81&0.75&0.22&0.91&0.87&0.35\\
       petet &0.99&0.92&4.34&0.79&0.79&0.12&0.85&0.85&0.32\\
        vm32 &0.99&0.92&0.99 &0.87&0.86&0.05&0.94&0.90&0.10\\ 
        wsj23 & 0.85& 0.79& 52.2& 0.64& 0.69&0.81 & -& -& -\\
        \end{tabular}
   \end{adjustbox}
        \caption{Parsing with 54GB RAM}
    \label{tab:RAM}

\end{table*}

\end{document}